# CORE BUILDING BLOCKS: NEXT GEN GEO SPATIAL GPT APPLICATION

**Ashley Fernandez**, **Swaraj Dube**

## ABSTRACT

This paper proposes MapGPT which is a novel approach that integrates the capabilities of language models, specifically large language models (LLMs), with spatial data processing techniques. This paper introduces MapGPT, which aims to bridge the gap between natural language understanding and spatial data analysis by highlighting the relevant core building blocks. By combining the strengths of LLMs and geospatial analysis, MapGPT enables more accurate and contextually aware responses to location-based queries. The proposed methodology highlights building LLMs on spatial and textual data, utilizing tokenization and vector representations specific to spatial information. The paper also explores the challenges associated with generating spatial vector representations. Furthermore, the study discusses the potential of computational capabilities within MapGPT, allowing users to perform geospatial computations and obtain visualized outputs. Overall, this research paper presents the building blocks and methodology of MapGPT, highlighting its potential to enhance spatial data understanding and generation in natural language processing applications.

*K***eywords** Geo-Spatial Analytics · Spatial Vector Representation · GPT · Embeddings

## 1 Introduction

Generative Pre-Trained (GPT)-based models have gained significant attention in recent years due to their ability to generate coherent and contextually relevant text. These models, based on the Transformer [1] architecture, have been successfully applied in various natural language processing tasks such as language translation [2-7], text summarization [8-12], and question answering [13-17]. The original GPT model and subsequent iterations, such as GPT-2 [18] and GPT-3 [19], have demonstrated remarkable performance in these domains, leveraging their large-scale pre-training on diverse text corpora.

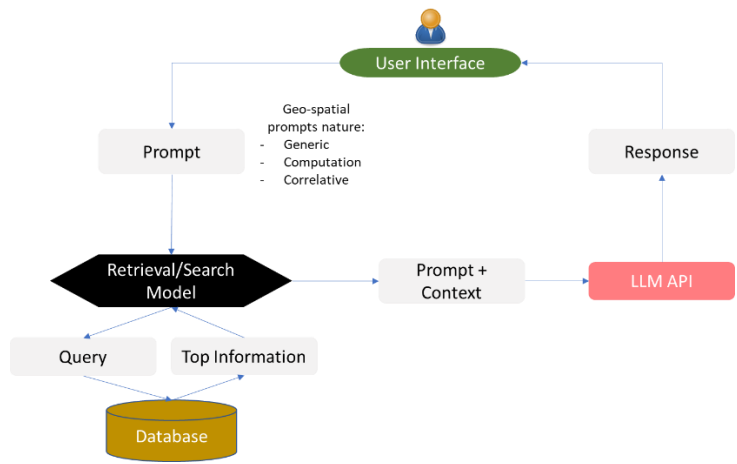

Figure 1: LLM Information Retrieval and Prompt Generation Workflow.

Figure 1 shows the workflow of how LLMs can generalize on localized knowledge via the information retrieval method. When a user query is received, the retrieval system searches for the most relevant documents or responses based on similarity measures or keyword matching. These methods leverage existing information and can provide more accurate and specific answers. Following this, LLMs generate answers by utilizing the information retrieved from various sources. They leverage their knowledge of language patterns, semantic relationships, and representations to generate relevant and coherent responses. This approach has both benefits and drawbacks which are listed and discussed below.

Primary benefits:

i. The potential to link LLMs with internal document databases, unlocking insights from internal data with LLM capabilities.
ii. The potential to have much more accurate and recent information.
iii. The resulting system could include references/citations to the original source documents from which the response was generated.

Primary drawbacks:

i. LLM retrieval models can still be inaccurate and hallucinate, providing out of context responses albeit typically less than when using LLMs without retrieval.
ii. Requires a strong information classification to mitigate privacy risks.
iii. Data leakage risk if LLM and search are not in the same infrastructure.

Secondary drawbacks:

It can be seen from Figure 1 that user prompts can have various natures i.e., the prompt may not also be generic prompts that are related to information retrieval. Prompts can also be computational in nature whereby the LLM model is expected to carry out computations with respect to the context of the prompt.

As mentioned that hallucinations and providing out of context answers are one of the main drawbacks of LLMs [20-23]. These can occur if robust vector representations of the information in the prompt are not present. In such cases, queries related to computation will also not be well understood because in this case, not only does the LLM need to understand the computational context of the prompt but also needs to carry out the computation. An example of a computation related query can be asking a prompt to compute the distances between 2 geospatial files. In this case, an LLM model will need to firstly understand the context of the question, accept files as input, understand that nature of the computation that must be carried out, generate code to execute the computation, optimize the code and the visualize the output. This is shown in Figure 2.

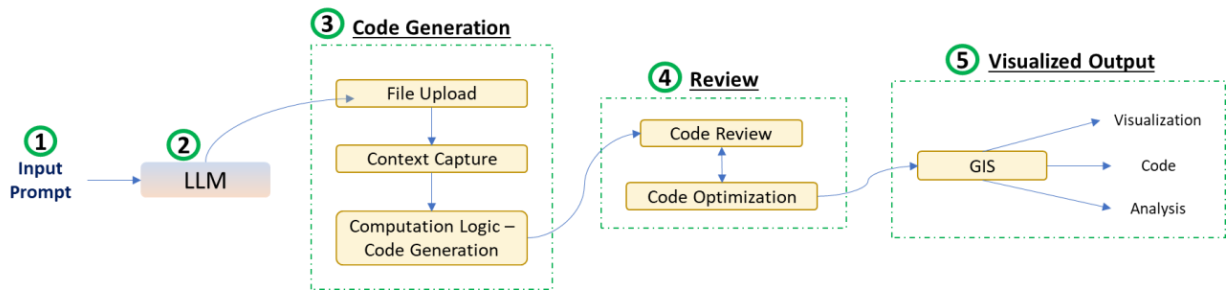

Figure 2: Computation Response Generation LLM Workflow.

To address the challenges mentioned above, Figure 2 shows the flow in which computational related queries can be handled. The first requirement is to have a geospatial database that contains robust vector representations can the understand prompt meaning especially in the computational context. Following this, users can upload files of interest on top which the computation needs to be performed. After the LLM understands the context of the question with the help of the vector representations in the geospatial database, a code generation step can be performed that generates code to perform the computation with respect to the query. Following this, code review can be performed by the user and the LLM could be prompted to carry out code optimization if needed. This output can then be integrated with GIS that provides visualized output of the computation, final code, and an analysis of the output.

While GPT-based models have shown impressive results in natural language processing, their application to spatial data has been limited [24-26]. For instance, when asked a location-specific question like "Where is the Coldplay event going to happen in Singapore?", a standard GPT model fails to grasp the vector representation of location data and provide an accurate response. The figure below shows an example demonstrating this using an LLM.

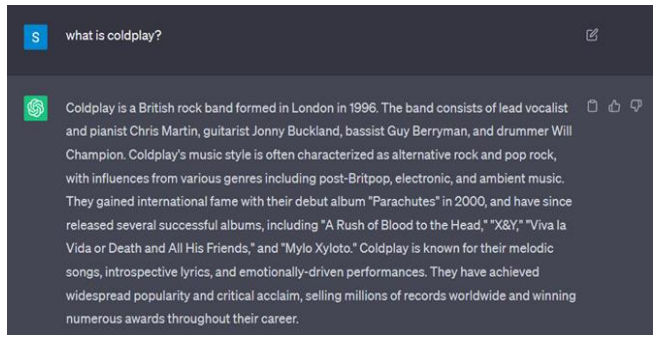

(a) LLM response to a generic prompt.

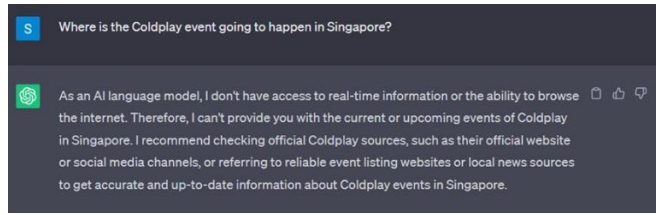

(b) LLM response to a location-based prompt.

Figure 3: LLM model response comparison between a generic prompt and a location-based prompt.

In Figure 3a), the LLM answered "What is Coldplay?" in a comprehensive manner but couldn't respond to a location-specific question about the Coldplay event in Singapore as shown in Figure 3b). GPT-based models rely on diverse vector representations to understand context, depending on the available training data. However, if vector representations for location-based data are missing, accurate responses may be challenging due to the lack of exposure during training.

**Solution**: Figure 3 demonstrates exactly the problem mentioned above which is a technical limitation of LLMs in itself. One potential solution of overcoming this limitation is via LLM retrieval capabilities that's integrated with external APIs for e.g., carrying out web search with GPT-based models. Figure 4 shows this workflow.

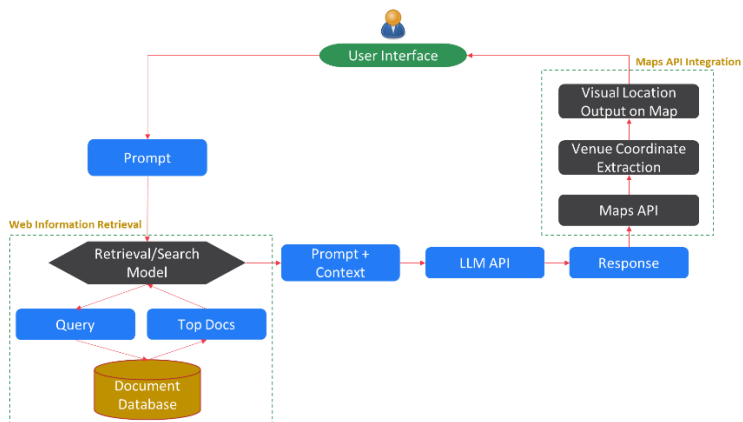

Figure 4: LLM model response comparison between a generic prompt and a location-based prompt.

Upon retrieval, any open-source LLM could be used to link it up to the web for information retrieval based on which the LLM could respond to location-based queries. For e.g., if the response is: "Coldplay event will take place in Marina Bay Sands", based on this response, the term 'Marina Bay Sands' can be extracted and passed into an API that provides the coordinates of this location in the form of latitude and longitude for instance. Using the coordinates, the location of the Coldplay event can finally be shown on the map in a visual format.

Though retrieval and API based solutions are common in the state-of-the-art, a challenge with this approach is that there are multiple dependencies on external APIs which could lead to 2 challenges:

  i. **Integration**: Complexity to package and maintain the entire system.
  ii. **Dependency risk**: If any API fails, the entire system could crash.

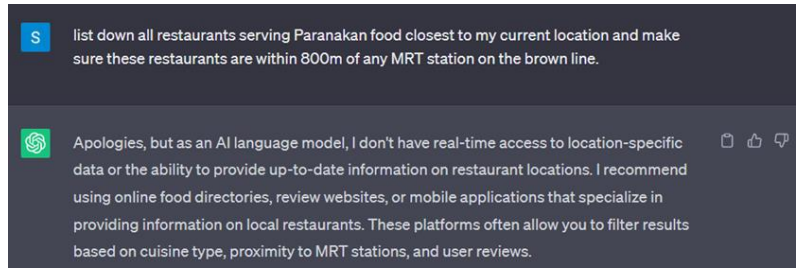

Figure 5: LLM model response comparison between a generic prompt and a location-based prompt.

In Figure 3, the LLM was able to understand about Coldplay since they are an internationally known music brand. However, if something is asked to the LLM completely in the context of Singapore especially containing location-based queries, the model again fails to provide a comprehensive answer as shown in Figure 5.

In the realm of spatial data processing, traditional approaches have primarily focused on specialized techniques such as geographic information systems (GIS), geospatial analysis, and map-based algorithms. These approaches have been instrumental in handling spatial data but often rely on structured and pre-processed datasets. The integration of GPT-based models with spatial data processing has been relatively unexplored, creating a significant gap in the literature.

This gap emphasizes the need for a specialized model like MapGPT that can effectively process and understand spatial data. By combining the strengths of GPT-based models with spatial data processing techniques, MapGPT aims to bridge the gap and enable more accurate and contextually aware responses to location-based queries. This paper presents an overview of the building blocks for a dedicated model like MapGPT to unlock the potential of spatial data in natural language understanding and generation.

## 2 Geo Spatial GPT Design Principles & Methodology

**Motivation**: The motivation behind the development of MapGPT lies in the recognition of the importance of spatial data and the need for AI models that can effectively process and understand such information. As mentioned that GPT-based models have primarily focused on textual data, neglecting the spatial aspect hindering their ability to provide accurate and meaningful responses to location-based queries. By introducing MapGPT, we aim to bridge this gap and empower AI models to comprehend and generate text in the context of spatial data. This also eliminates dependencies on map-based APIs and opens up new possibilities for enhancing location-based services, navigation systems, and various other applications that heavily rely on the interplay between language and location using one model.

## 2.1. Model Architecture

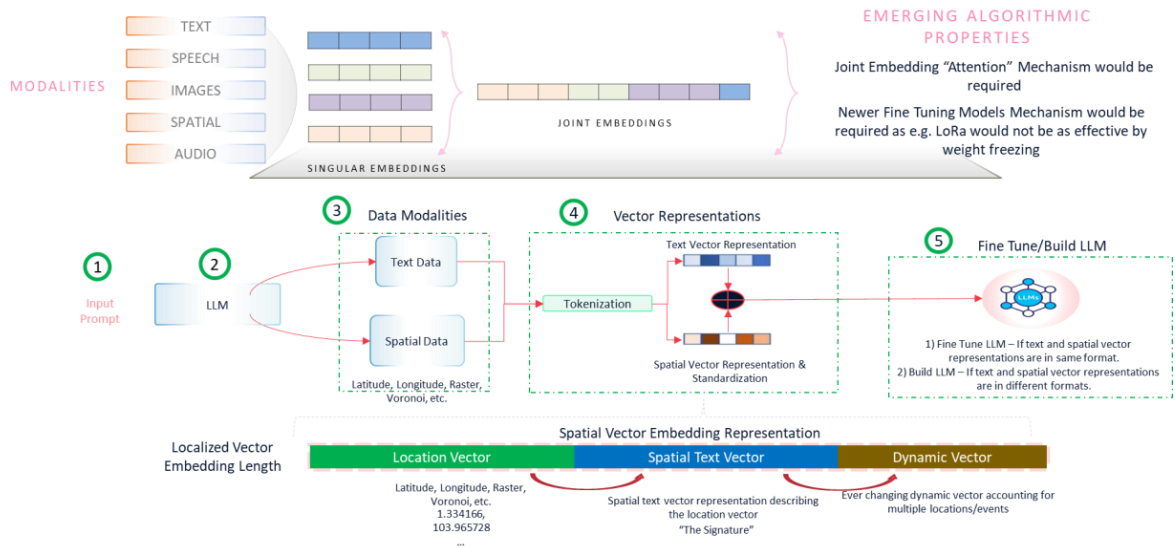

Figure 6: Proposed Technical Solution Architecture of MapGPT.

At a data modality level, location-based queries will contain text data and spatial data for e.g., "Where is the best curry rice shop in Singapore?". MapGPT leverages the power of both text and spatial data to enable comprehensive and contextually aware responses in the realm of location-based information. This fusion of data sources enables more accurate and meaningful responses to queries related to maps, directions, points of interest, and other location-based information.

Tokenization is a fundamental step in the MapGPT training process [27-30], where the input text and spatial data are divided into individual units called tokens. These tokens can be words, phrases, or even subword units, depending on the specific tokenization strategy used which is mentioned later during the paper. The purpose of tokenization is to break down the input data into meaningful units that can be processed by the model.

By segmenting the query into tokens, MapGPT gains a granular understanding of the input, capturing the syntactic and semantic structure of the text and the spatial attributes of the data. This process enables MapGPT to handle and process complex and diverse inputs, facilitating its ability to generate accurate and contextually appropriate responses and insights related to location-based queries.

After the tokenization process, MapGPT generates vector representations for both textual and spatial data. The text vector representation captures the semantic meaning and contextual information of the input text, allowing MapGPT to understand and generate responses based on its training on a vast corpus of text data. On the other hand, the spatial vector representation encodes the spatial attributes, such as coordinates, distances, and relationships between locations, enabling MapGPT to comprehend and process spatial data effectively.

The spatial representation holds the key to unlocking MapGPT's potential. This is a novel vector representation we are proposing that is a combination of 3 vector representations for which the breakdown is as follows:

i. **Location Vector (fixed representation)**: The location vector is a vector representation that encodes spatial attributes associated with a particular location that includes information such as latitude, longitude, and other relevant spatial coordinates. The location vector serves as a compact representation of the geographic position, allowing MapGPT to understand the specific location being referred to in a text or query.

ii. **Spatial Text Vector (fixed representation)**: This vector contains the representation that describes the location vector. It contains the representation that describes the location vector in the context of the associated text. This vector captures the relationship between the text and the spatial attributes, enabling MapGPT to link the textual information with the specific location described by the location vector.

iii. **Dynamic Vector (dynamic representation)**: Location and spatial text vectors will have fixed representations, however, the dynamic vector represents granular level information pertaining to locations/events happening within the main location for e.g., if the main building is named "The Signature", there could be several events that could happen in the same building such as "Partner Summit" happening the next month but after 6 months, another event could take place in the same building, hence the need for a dynamic vector represent to the cater for such scenarios.

Open source LLMs typically uses a standardized text representation for training. However, in this case, we are proposing a novel spatial vector representation which is a combination of 3 vector representations. For this reason, a vector representation standardization process must be carried out for the spatial vector representation in way such that its format is similar to the text vector representation upon which the fine-tuning of LLMs can be carried out. By combining these two types of vector representations (text and spatial vector representations), MapGPT achieves a comprehensive understanding of both text and spatial aspects, enabling it to provide accurate and contextually relevant information in response to location-based queries.

In situations where the length of the spatial vector representation differs from the standard text vector representation, it may be necessary to train a custom GPT model. This process involves building a new model architecture that can accommodate the combined text and spatial data. The training process requires a dataset that includes both textual and spatial information, which is used to fine-tune the model specifically for the task at hand. By carrying out training, we can ensure that it effectively integrates spatial data and produces accurate and contextually aware responses to location-based queries.

**2.2. Geospatial, GPT Vector Form Database Design and Management**

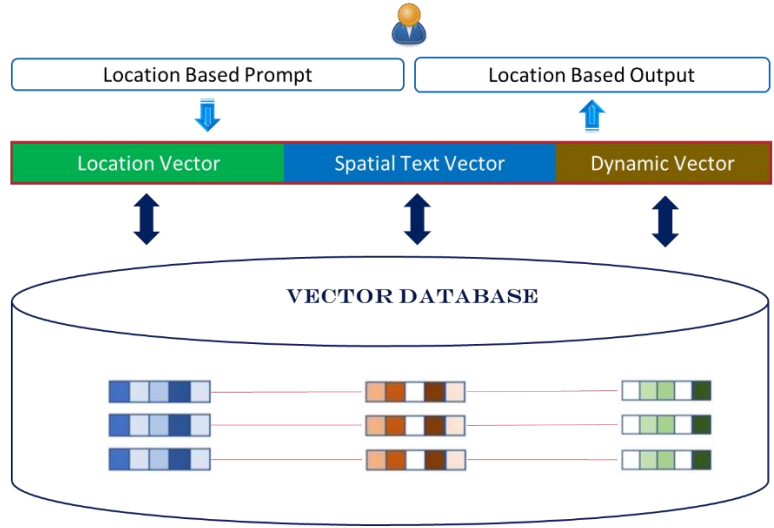

Figure 7: MapGPT post-training usage.

Figure 7 shows the underlying vector database via which responses to prompts can be provided. Vector database in this case stores high dimensional vector representations of text and spatial data. When a user enters a location-based prompt to get a response, the methodology involves leveraging the stored text vector representation and spatial text vector representation in the vector database. This vector database contains pre-computed embeddings for various textual and spatial elements. When a user prompt is received, the prompt is tokenized and converted into a text vector representation. Simultaneously, the spatial information in the prompt is extracted and transformed into a spatial text vector representation.

The system then performs a search in the vector database using the text vector representation and the spatial text vector representation. This search matches the user's input with the most relevant vector representations stored in the database. The retrieved context contains the encoded information necessary to provide a contextually aware response to the user's location-based prompt. The retrieved context will be in raw form, hence, to ensure that the responses to the location-based prompts are natural, a technique known as retrieval-augmented-generation (RAG) will be carried out whereby, the retrieved context will be fed into the LLM along with the prompt which will enable human like responses with the help of underlying knowledge of LLMs and the retrieved context.

The system then processes the retrieved vectors and generates a response that combines both textual and spatial information. This response is tailored to the user's query and provides relevant and accurate information based on the location-based context. By leveraging the stored vector representations in the databases, MapGPT ensures efficient and effective retrieval of information for location-based queries.

**2.3. Considerations for Text Representation Scheme Selection**

There are various techniques for choosing a suitable text representation scheme for map data. It explores the different considerations in representing geographical information, such as coordinates, addresses, and landmarks, as text tokens. It discusses the trade-offs between informativeness, compactness, and compatibility with the GPT model's architecture. Examples of representation schemes, including tokenization approaches. Procedural steps and key considerations are provided to guide the implementation of the text representation scheme:

   a) **Considerations for Text Representation Scheme Selection**: When selecting a text representation scheme, several considerations must be taken in consideration. These include the level of detail required, the specific spatial attributes to be represented, and the desired level of semantic understanding. Additionally, factors such as vocabulary size, representation compactness, and compatibility with the GPT model's architecture should be considered.

   b) **Tokenization Approaches**: Tokenization is a fundamental aspect of representing map data as text tokens. Several tokenization approaches can be employed, depending on the specific requirements of the MapGPT model. Here are a few examples:

   I. **N-gram Tokenization**: N-gram tokenization involves dividing the map data into contiguous sequences of N tokens. For instance, a bi-gram tokenization of the phrase "New York City" would result in tokens like "New York" and "York City." This approach provides context by considering adjacent tokens.

   II. **Subword Tokenization**: Subword tokenization splits words into smaller subword units. This technique is useful for handling out-of-vocabulary words and reducing vocabulary size. Methods such as Byte Pair Encoding (BPE) or SentencePiece can be employed for subword tokenization.

   III. **Semantic Tokenization**: Semantic tokenization involves assigning tokens based on the semantic meaning of the underlying data. For example, in a map context, tokens like "city," "street," or "landmark" can be used to represent different types of locations. This approach enables the model to understand the spatial context more explicitly.

Example of Representation Schemes: To illustrate the selection process, let's consider two scenarios:

   a) **Scenario 1**: Representing Coordinates In this case, converting geographical coordinates (latitude and longitude) into a text representation is crucial. One approach could be to represent coordinates as readable textual descriptions. For example, the coordinate (1.3008° N, 103.9122° E) would be represented as "Singapore City coordinates." Alternatively, incorporating coordinates into predefined spatial templates like "[Latitude], [Longitude]" would maintain the original information while allowing flexibility in processing.

   b) **Scenario 2**: Representing Addresses representation requires parsing and normalizing unstructured address strings. Techniques such as rule-based parsing or statistical parsing can be employed to extract structured information. For example, the address "123 East Coast, Singapore City " could be parsed into tokens like "street_number: 123," "street_name: East Coast," and "city: Singapore."

Normalization ensures consistent address formats and resolves variations, such as abbreviations or alternative spellings. Hence summing up the key considerations in the selection process, while choosing a text representation scheme, several considerations should be kept in mind:

   a) **Informativeness vs. Compactness**: Striking a balance between providing informative representations and maintaining compactness is essential. Detailed representations may improve understanding but may also increase computational complexity and memory requirements.

   b) **Compatibility with GPT Model Architecture**: The chosen text representation scheme should align with the GPT model's architecture to facilitate effective training and inference. Consider how the model's attention mechanism can leverage the representations for accurate predictions.

**2.4. Landmark Description Retrieval Strategies**

Retrieving accurate and contextually relevant descriptions of landmarks is crucial for enhancing MapGPT's understanding and generation capabilities. This section explores various methods for obtaining textual descriptions of landmarks. Techniques such as leveraging geospatial databases, utilizing online resources, and utilizing specialized

APIs are discussed. Additionally, considerations for relevance scoring, filtering, and quality assurance are addressed to ensure the retrieval of valuable landmark descriptions for the MapGPT model.

1) **Leveraging Geospatial Databases**: Geospatial databases provide a vast amount of information about landmarks. By leveraging these databases, relevant textual descriptions can be obtained [31,32]. Techniques for utilizing geospatial databases include:

   a) **Querying by Location**: Geospatial databases often provide search capabilities based on location coordinates or addresses. By querying the database with the location information, relevant descriptions of nearby landmarks can be retrieved. This approach ensures that the generated text is closely associated with the specific location.

   b) **Extracting Descriptions from Attributes**: Geospatial databases typically contain additional attributes associated with landmarks, such as names, categories, and descriptions. Extracting textual descriptions from these attributes can provide valuable information about landmarks. Techniques such as text extraction and natural language processing can be used to extract relevant descriptions.

2) **Utilizing Online Resources**: Online resources such as travel websites, tourism portals, and review platforms offer a wealth of information about landmarks [33-35]. These resources can be leveraged to retrieve textual descriptions. Methods for utilizing online resources include:

   a) **Web Scraping**: Web scraping techniques can be employed to extract relevant descriptions from websites. By crawling travel websites or tourism portals, textual content associated with landmarks, such as descriptions, historical information, and visitor reviews, can be extracted. However, it is important to respect a website's terms of service and be mindful of legal and ethical considerations.

   b) **Natural Language Processing on Reviews**: Online review platforms often contain detailed user-generated descriptions of landmarks. By applying natural language processing techniques, sentiment analysis, and topic modeling, relevant descriptions can be extracted from user reviews. This approach provides valuable insights into the subjective experiences and opinions associated with landmarks.

3) **Geo based APIs**: Specialized APIs designed specifically for accessing geographical information can be utilized to retrieve landmark descriptions. These APIs offer structured data and descriptions associated with landmarks. Techniques for utilizing specialized APIs include:

   a) **Places API**: Platforms such as Google Maps provide APIs that offer access to comprehensive data about places and landmarks. By querying the Places API with location information, textual descriptions, categories, and other relevant attributes can be obtained. This approach ensures the retrieval of accurate and up-to-date information.

   b) **Knowledge Graph APIs**: Knowledge graph APIs, such as Wikipedia or Wikidata, provide structured and detailed information about various landmarks. By querying these APIs with specific location or landmark identifiers, textual descriptions, historical facts, and associated information can be retrieved. This approach enhances the model's understanding of the cultural, historical, and geographical aspects of landmarks.

4) **Considerations for Retrieval Methods:** When retrieving landmark descriptions, several considerations should be considered:

   a) **Relevance Scoring**: Implementing a relevance scoring mechanism can help filter and rank descriptions based on their relevance to the given location or context. Techniques such as keyword matching, semantic similarity, or machine learning models can be employed to assign relevance scores to the retrieved descriptions.

   b) **Filtering and Quality Assurance**: Ensuring the quality and accuracy of retrieved descriptions is crucial. Applying filters based on factors such as source credibility, language quality, and redundancy can help eliminate unreliable or low-quality descriptions.

**2.5. Enhancing MapGPT with Landmark Descriptions**

This involves the process of encoding spatial relationships in text form which is essential for the MapGPT model to understand the relative positions and distances between locations accurately. This section explores various strategies for representing spatial relationships using textual markers or tokens. Techniques such as descriptive phrases, directional indicators, proximity indicators, and hierarchical representations are discussed. Considerations for handling complex spatial relationships, ensuring compatibility with the GPT model's attention mechanism, and maintaining coherence in the encoded text are addressed.

  a) **Descriptive Phrases**: Descriptive phrases provide explicit information about the spatial relationships between locations. For example, using phrases like "north of," "nearby," "adjacent to," or "opposite" can convey the relative positions effectively. These phrases can be incorporated into the text representation scheme as specific tokens or as part of the landmark descriptions.

  b) **Directional Indicators**: Directional indicators offer information about cardinal directions or orientations. Tokens such as "north," "south," "east," and "west" can be used to denote the direction of one location from another. Additionally, tokens representing relative angles, such as "45 degrees clockwise," can be employed for more precise spatial relationships.

  c) **Proximity Indicators**: Proximity indicators provide information about the proximity or distance between locations. Terms such as "close to," "far from," or "within a radius of" can be used to express spatial relationships based on distance. Incorporating these indicators into the text representation scheme allows the MapGPT model to understand the relative proximity of different locations.

  d) **Hierarchical Representations**: Hierarchical representations enable the encoding of complex spatial relationships involving multiple locations. Hierarchical structures can be used to represent relationships such as containment, hierarchy, or grouping. For example, a hierarchical representation can capture relationships like "the city is within the county, which is within the state." These hierarchical relationships can be encoded using specific tokens or by defining rules within the text representation scheme.

Considerations for Encoding Spatial Relationships: When encoding spatial relationships in text form, the following considerations should be considered:

  a) **Coherence and Natural Language Flow**: The encoded text should maintain coherence and natural language flow. The spatial relationship markers or tokens should be seamlessly integrated into the text, ensuring that the resulting text reads naturally and is easy to understand.

  b) **Compatibility with GPT Model's Attention Mechanism**: The chosen encoding scheme should be compatible with the attention mechanism of the GPT model. Attention mechanisms allow the model to focus on relevant spatial relationships during training and inference, enabling accurate predictions and generation.

  c) **Handling Complex Spatial Relationships**: Complex spatial relationships, such as overlapping areas or non-linear configurations, require careful consideration. Techniques like using additional descriptive phrases, introducing higher-level abstractions, or incorporating geometric representations can aid in encoding these complex relationships effectively.

# 3 EVALS (Validating & Iterative Improvement)

Validation and iterative improvement are crucial steps in the development and refinement of the MapGPT model. The validation process involves evaluating the model's performance and assessing its ability to accurately handle location-based queries and provide meaningful responses. This phase involves the following technical considerations:

  i. **Geospatial Metrics**: Geospatial-specific metrics are employed to evaluate the model's performance. These metrics include spatial accuracy, geocoding precision, and contextual relevance, which are vital for maps-related tasks.

  ii. **Specialized Datasets**: Datasets specifically curated for maps-related applications are utilized for validation. These datasets encompass a wide range of geographical queries, from route planning and point-of-interest searches to geocoding and spatial analysis.

  iii. **Cross-Domain Evaluation:** The model is assessed for its versatility across various maps-related domains, including navigation, location-based recommendations, geographic information retrieval, and spatial analysis.

Iterative improvement is then carried out based on the insights gained from the validation process. This involves analyzing the model's shortcomings, identifying areas of improvement, and implementing necessary changes. The iterative improvement process may include fine-tuning the model's parameters, adjusting the training data, refining the tokenization and vector representation methods, and incorporating feedback from users and domain experts. This phase involves the following technical considerations:

  i. **Geospatial Fine-tuning**: In-depth fine-tuning of the model's parameters is performed, with a focus on geospatial context. This includes optimizing geocoding accuracy, enhancing geographic knowledge, and refining location-specific response generation.

  ii. **Geospatial Data Augmentation**: The model's geospatial knowledge is expanded through the incorporation of diverse geospatial datasets. These datasets cover a wide spectrum of maps-related scenarios, enabling MapGPT to provide more comprehensive and accurate information.

  iii. **Advanced Geospatial Tokenization**: Specialized geospatial tokenization techniques are explored to improve the model's understanding of location-related queries. This includes adapting tokenizers to handle geocoordinates, addresses, place names, and other geospatial data effectively.

  iv. **User and Expert Feedback Integration**: Valuable insights from user feedback and domain experts are systematically integrated into the model's training and refinement process. This iterative feedback loop ensures that MapGPT continually adapts to evolving user needs and maintains its relevance in maps-related applications.

## 4 Practical Applications of the MapGPT model

As this paper is themed as an emerging concept for maps, the following are some of the practical applications of the MapGPT model:

  i. **Spatial Data Analysis**: MapGPT enables the analysis of spatial data by understanding and processing natural language queries related to spatial information. This can be applied in various domains, including urban planning, transportation management, and environmental monitoring.

  ii. **Geospatial Computation**: With its computational capabilities, MapGPT allows users to perform geospatial computations, such as measuring distances between locations, calculating travel times, or determining optimal routes based on geospatial input files.

  iii. **Spatial Data Visualization**: MapGPT can generate visual representations of spatial data, allowing users to explore and understand complex spatial patterns, visualize geospatial computations, and gain insights from the data.

  iv. **Natural Language Interfaces for Maps**: MapGPT can serve as a natural language interface for maps, allowing users to interact with maps and spatial data through conversational queries, making it more accessible and intuitive for users to retrieve information.

  v. **Context-Aware Location-Based Services**: MapGPT enhances the context-awareness of location-based services, enabling them to understand user intent and provide tailored recommendations based on the user's current location, preferences, and contextual information.

MapGPT can have a wide range of practical applications in the field of spatial data analysis and geospatial computation. It enables the analysis and processing of spatial data through natural language queries providing a more interactive experience when dealing with maps. With its computational capabilities, MapGPT can facilitate geospatial computations such as distance measurement, travel time calculation, and route optimization. Additionally, MapGPT can support spatial data visualization, allowing users to explore complex spatial patterns and visualize geospatial computations. As a natural language interface for maps, MapGPT enhances user interaction with maps and spatial data, making it more accessible and intuitive. Moreover, its context-awareness enhances location-based services by understanding user intent and providing tailored recommendations based on the user's current location and preferences.

This is primarily done through data organization whereby each geo data will be hosted in a graph database whereby nodes and edges represent relationships. This will help in retrieving the appropriate context with respect to the user prompt thus enabling more accurate responses. Once the right context has been retrieved, it must be explained in a comprehensive human like manner to enrich user experience, this is where LLMs come in whereby the user prompt and the retrieved context will be input into the LLM after which the responses will be generated.

## 5  Conclusion

In this paper, we have presented MapGPT, an approach that combines large language models (LLMs) with spatial data processing techniques to improve natural language understanding and spatial data analysis. MapGPT leverages the strengths of LLMs and geospatial analysis to provide more accurate and contextually aware responses to location-based queries. By building LLMs on spatial and textual data and employing custom tokenization and vector representations for spatial information, MapGPT enhances the integration of language models with spatial data. Additionally, we have explored the potential of computational capabilities within MapGPT, enabling users to perform geospatial computations and visualize the outputs. Through our findings, we have demonstrated the importance and applicability of MapGPT in advancing the field of natural language processing and spatial data analysis. Future work can focus on refining the spatial vector representations and further enhancing the computational capabilities of MapGPT to address the evolving challenges in spatial data processing.